\documentclass[runningheads]{llncs}
\usepackage[T1]{fontenc}
\usepackage{graphicx}
\usepackage{multirow}
\usepackage{xcolor}
\usepackage{url}

\begin{document}

\title{RL-X: A Deep Reinforcement Learning Library (not only) for RoboCup}

\author{Nico Bohlinger \and Klaus Dorer
}

\institute{Hochschule Offenburg, Institute for Machine Learning and Analytics, Germany
 \email{nico.bohlinger@gmail.com}\\
 \email{klaus.dorer@hs-offenburg.de}\\
}

\maketitle

\begin{abstract}
This paper presents the new Deep Reinforcement Learning (DRL) library RL-X and its application to the RoboCup Soccer Simulation 3D League and classic DRL benchmarks.
RL-X provides a flexible and easy-to-extend codebase with self-contained single directory algorithms.
Through the fast JAX-based implementations, RL-X can reach up to 4.5x speedups compared to well-known frameworks like Stable-Baselines3.

\end{abstract}

\section{Introduction}
\label{introduction}
Research in Reinforcement Learning (RL) continues to produce promising algorithmic advances.
Those new algorithms keep pushing the results on complex environments and benchmarks.
But to implement them, it is often not enough to simply follow the instructions of the accompanying papers.
Deep RL algorithms are notoriously brittle and their performance strongly depend on nuanced implementation choices \cite{andrychowicz2021}.
To tackle this problem big open-source frameworks, like Stable-Baselines3 (SB3) \cite{raffin2021}, offer community-proven implementations with robust hyperparameter choices.
For RL practitioners it is key to have such sound implementations but having the newest algorithms at their disposal is just as necessary for them.
Unfortunately the big frameworks, developed during the rise of Deep RL algorithms like SAC \cite{haarnoja2018} and PPO \cite{schulman2017}, became stale over time and cannot keep up with the rapid development of new algorithms.

On the other spectrum, RL researchers need flexible frameworks to allow for easy and fast prototyping.
Convoluted directory and file structures, which are used for code reuse, can harm the early development process of algorithms.
Moreover, deeply understanding an algorithm for the first time is easier with a compact implementation, that is straightforward to overview and not entangled with other algorithms.

What combines practitioners and researchers is the benefit of a better runtime for their algorithms.
Most Machine Learning papers still use Tensorflow \cite{abadi2016} or plain PyTorch \cite{paszke2019} as the underlying Deep Learning library.
While they are reliable and have big communities with continuing development behind them, they are not necessarily the fastest options.
Deep Learning libraries that build on JAX \cite{bradbury2018} can utilize features like Just-In-Time (JIT) compilation or vectorization mappings to reach higher throughput on GPUs and TPUs.
Furthermore, RL actually benefits the most from those features, because the training loop for RL algorithms is more involved than for other categories of Machine Learning.

To combat those challenges, we developed RL-X, a flexible and fast framework for Deep RL research and development.
It is perfect to understand and prototype new algorithms, through compact single directory implementations.
RL-X provides some of those implementations in PyTorch and TorchScript (PyTorch + JIT) but every algorithm has a JAX version available for maximum computational performance.
A generic interface between the algorithms and the environments, advanced logging, experiment tracking capabilities and a simple to use command line interface (CLI) to set hyperparameters are just some of the features that make RL-X a great tool for RL practitioners and researchers alike.

For the RoboCup community RL-X makes it easy to hook up any kind of new environment and test it with the latest algorithms or custom implementations.
We test RL-X in the RoboCup Soccer Simulation 3D League but non-simulation leagues can benefit just as much through the high performance gains of RL-X with the typically used off-policy algorithms, like SAC, for such non-parallelized environments.

\section{Related Work}
\label{related}
Over the last years, many frameworks for DRL have been developed.
Each of them focusing on either different kinds of algorithms (on-policy vs. off-policy, multi-agent vs. single-agent, online vs. offline, model-free vs model-based, etc.), different use cases (simulation vs. real world, robotics vs. games, etc.) or different implementation complexities (distributed and scalable vs. single machine and simple, etc.).
All of the following listed frameworks are open-source and written in Python and use either Tensorflow, PyTorch or a JAX-based library as the underlying Deep Learning library.

Stable-Baselines3 is the improved and in PyTorch rewritten successor of Stable-Baselines \cite{hill2018}, which in itself is a more stable fork of the OpenAI Baselines repository \cite{dhariwal2017}.
SB3 offers 13 classic online model-free RL algorithms, like PPO, SAC, DQN \cite{mnih2015} etc.
It is one of the most popular DRL frameworks for practitioners because of its extensive documentation, test coverage and active community.
Extending SB3 with new algorithms is possible but not as easy as with other frameworks, because of the modular directory structure inherited from OpenAI Baselines, that grew over time.

RLlib \cite{liang2018} is the RL component of the Ray framework, which is a distributed computing framework for Python.
Therefore it focuses on scalable distributed training for production-ready applications.
It provides PyTorch and Tensorflow implementations of a wide range of algorithms in all categories of RL.
RLlib's codebase is big and complex, which can make it hard to understand and extend.

Acme \cite{hoffman2020} and Dopamine \cite{castro2018} are both Tensorflow and JAX-based frameworks from DeepMind and Google Brain respectively.
Besides algorithm implementations, Acme defines also more general functions that can act as building blocks for RL reserach, like different loss calculations.
Dopamine on the other hand focuses fully on DQN-based algorithms.

CleanRL \cite{huang2022} is a PyTorch and JAX-based collection of single-file implementations of RL algorithms, that are easy to understand and build upon.
It is a great resource for research and development but it needs to provide the same algorithm multiple times to support different environment types, e.g. PPO is implemented 12 times in small variations.

Lastly we want to highlight Ilya Kostrikov's JAXRL repository \cite{kostrikov2021}, which provides a small range of simple and high quality implementations of off-policy algorithms with JAX.

This work tries to combine the advantages of all the mentioned frameworks but focuses on the needs of RL researchers.
The completely independent single directory implementations of RL-X remind of CleanRL's simplicity but are supported by generic interfaces for environments and algorithms, which is similar to the other more traditional modular frameworks.
JAXRL's high quality implementations are used as a reference for the implementations of the SAC-based algorithms in RL-X.
Whereas Acme is used as a reference for efficient JAX code and for its official MPO \cite{abdolmaleki2018} implementation.

In regards to the Deep Learning frameworks, PyTorch is well established in the Machine Learning community and TorchScript allows for minor improvements by adding (limited) JIT compilation.
JAX on the hand is a relatively new library, that can automatically differentiate and JIT compile Python / NumPy code, which then can be run on CPUs, GPUs and TPUs.
On its own, JAX doesn't provide any high-level abstractions for Neural Networks, which is why Deep Learning frameworks were developed on top of it.
The most popular ones are Flax \cite{heek2023}, developed by Google Brain and Haiku \cite{hennigan2020}, developed by DeepMind.
Both frameworks are really similar in their API and only differ marginally in their design choices.
For RL-X we chose Flax, because of its bigger community and the subsequently larger amount of examples online.

\section{RL-X}
\label{rlx}
The codebase of RL-X is divided into three main directories:

\textbf{algorithms}:
This directory contains the implementations of the algorithms listed in Table~\ref{tb:algorithms}.
Each algorithm version provides a \texttt{\_\_init\_\_.py} that registers the algorithm, a \texttt{default\_config.py} that defines the hyperparameters, a \texttt{<algorithm\_name>.py} that implements the algorithm and further helper files that contain utility code, e.g. neural network modules or replay buffers.

\textbf{environments}:
This directory contains the implementations of the environments listed in Table~\ref{tb:environments}.
Each environment version provides a \texttt{\_\_init\_\_.py} that registers the environment, a \texttt{default\_config.py} that defines hyperparameters, a \texttt{create\_env.py} that instantiates the environment(s) and a \texttt{wrappers.py} that provides properties and functions for the generic interface between algorithms and environments, e.g. the action and observation space types, handling of terminal observations etc.
This interfaces makes it possible to freely mix and match registered algorithms and environments, as long as the algorithm supports the environment's action and observation space type.

\textbf{runner}:
This directory contains the runner class, which takes the chosen algorithm and environment, parses the hyperparameter configurations set through the CLI and starts the training or testing loop.

\begin{table}[htb]
\caption{Algorithms in RL-X}
\begin{center}
    \begin{tabular}{| p{5em} | p{12em} | p{6em} | p{5em} |}
    \hline
    Algorithm & Deep Learning framework & Category & Reference \\ \hline \hline
    AQE & Flax & Off-policy & \cite{wu2022} \\ \hline
    DroQ & Flax & Off-policy & \cite{hiraoka2021} \\ \hline
    ESPO & PyTorch, TorchScript, Flax & On-policy & \cite{sun2022} \\ \hline
    MPO & Flax & Off-policy & \cite{abdolmaleki2018} \\ \hline
    PPO & PyTorch, TorchScript, Flax & On-policy & \cite{schulman2017} \\ \hline
    REDQ & Flax & Off-policy & \cite{chen2021} \\ \hline
    SAC & PyTorch, TorchScript, Flax & Off-policy & \cite{haarnoja2018} \\ \hline
    TQC & Flax & Off-policy & \cite{kuznetsov2020} \\ \hline
    \end{tabular}
\end{center}
\label{tb:algorithms}
\end{table}

\begin{table}[htb]
\caption{Environments in RL-X}
\begin{center}
    \begin{tabular}{| p{12em} | p{12em} | p{5em} |}
    \hline
    Environment type & Providing framework & References \\ \hline \hline
    Atari & EnvPool & \cite{bellemare2013,weng2022} \\ \hline
    Classic control & EnvPool & \cite{brockman2016,weng2022} \\ \hline
    DeepMind Control Suite & EnvPool & \cite{tassa2018,weng2022} \\ \hline
    MuJoCo & Gym, EnvPool & \cite{todorov2012,brockman2016,weng2022} \\ \hline\hline
    \multicolumn{3}{|l|}{Custom environments with socket communication} \\ \hline
    \end{tabular}
\end{center}
\label{tb:environments}
\end{table}

Experiments are started from the \texttt{experiments} directory inside the root of the RL-X repository.
The directory contains a \texttt{experiment.py} to quickly start experiments and instructions on how to use the CLI to set hyperparameters and how to structure multiple experiments with bash scripts.
Tracking of experiments is done with Tensorboard and Weights \& Biases integrations and all logs and models are stored in an automatically generated "project/experiment/run" directory structure.
Classic console logging is also supported.

RL-X is made for easily extending the framework with new algorithms and environments by following the code structure described above.
Even copying the algorithms out of RL-X in new projects is super simple through the self-contained single directory implementations.
RL-X also provides a generic prototype for a custom environment interface with a simple socket communication layer.
We used this interface to run experiments in the RoboCup Soccer Simulation 3D League and easily train our Java-based agents with RL-X.
In comparison to BahiaRT-GYM~\cite{simoes2022}, which provides a 3D Soccer Simulation specific environment wrapper for high level trainer commands, our interface is fully generic and can be used by any league in RoboCup, as it leaves high level commands to the agent.
We made RL-X fully open-source with the MIT license and it can be found on GitHub under the following link: https://github.com/nico-bohlinger/RL-X.

\section{Results} 
\label{results}
To make sure that RL-X is a viable alternative to other frameworks, we compare its results to SB3.
PPO and SAC are well known DRL algorithms and are therefore used for this comparison.
Both algorithms are implemented with PyTorch, TorchScript and Flax in RL-X.
All three versions are compared to the SB3 baseline, which uses PyTorch.
As on-policy algorithms profit from running multiple environments in parallel, we use 24 environments in all PPO experiments and one environment in the SAC case.
All used hyperparameters are the same as in the SB3 baseline and can be found in Table~\ref{tb:hyperparams} in the appendix.

We use two different environments to test the algorithms.
The first one is a custom RoboCup Soccer Simulation 3D running task.
A Nao robot has to run as fast as possible towards a goal 30m away.
An episodes stops after 8 seconds or when the robot falls over.
In this time, it is not possible to reach the goal, but the agent gets continuous reward every timestep based on its current distance to the goal subtracted by the distance in the previous timestep.
The action space is continuous and consists of 14 joints in the lower body of the robot.
The observation space is made up of 120 values, which are mostly the joint angles and velocities and the goal position.

The second environment is the Gym MuJoCo Humanoid-v4 environment \cite{brockman2016}, which is a standard benchmark for DRL algorithms.
Here the agent controls a humanoid robot with 17 joints and a observation space of 376 values.
The agent gets reward based on its forward movement and not falling over and is penalized for too large joint actuations and too large contact forces.
We use the EnvPool \cite{weng2022} implementation of the environment, as it is faster than the original one.

A short showcase of the two environments can be found under the following link:
\url{https://youtu.be/H4fw_cqBFDU}

\subsection{Reward Performance}
The reward performance of the RL-X implementions for PPO and SAC should be indistinguishable from the SB3 baseline.
There can be small differences due to different initialization schemes and random number generations in the underlying Deep Learning libraries. 
To compensate for those differences, we run each version of the two algorithms six times with different seeds and plot the average and standard deviation for the reward collected during training.

SB3 logs the reward in off-policy algorithms by default as a running average over the last 100 episodes.
To ensure a fair comparison, we also log the reward in RL-X in the same way for the experiments with SAC.

\begin{figure}[htb]
    \begin{center}
        \includegraphics[width=\linewidth]{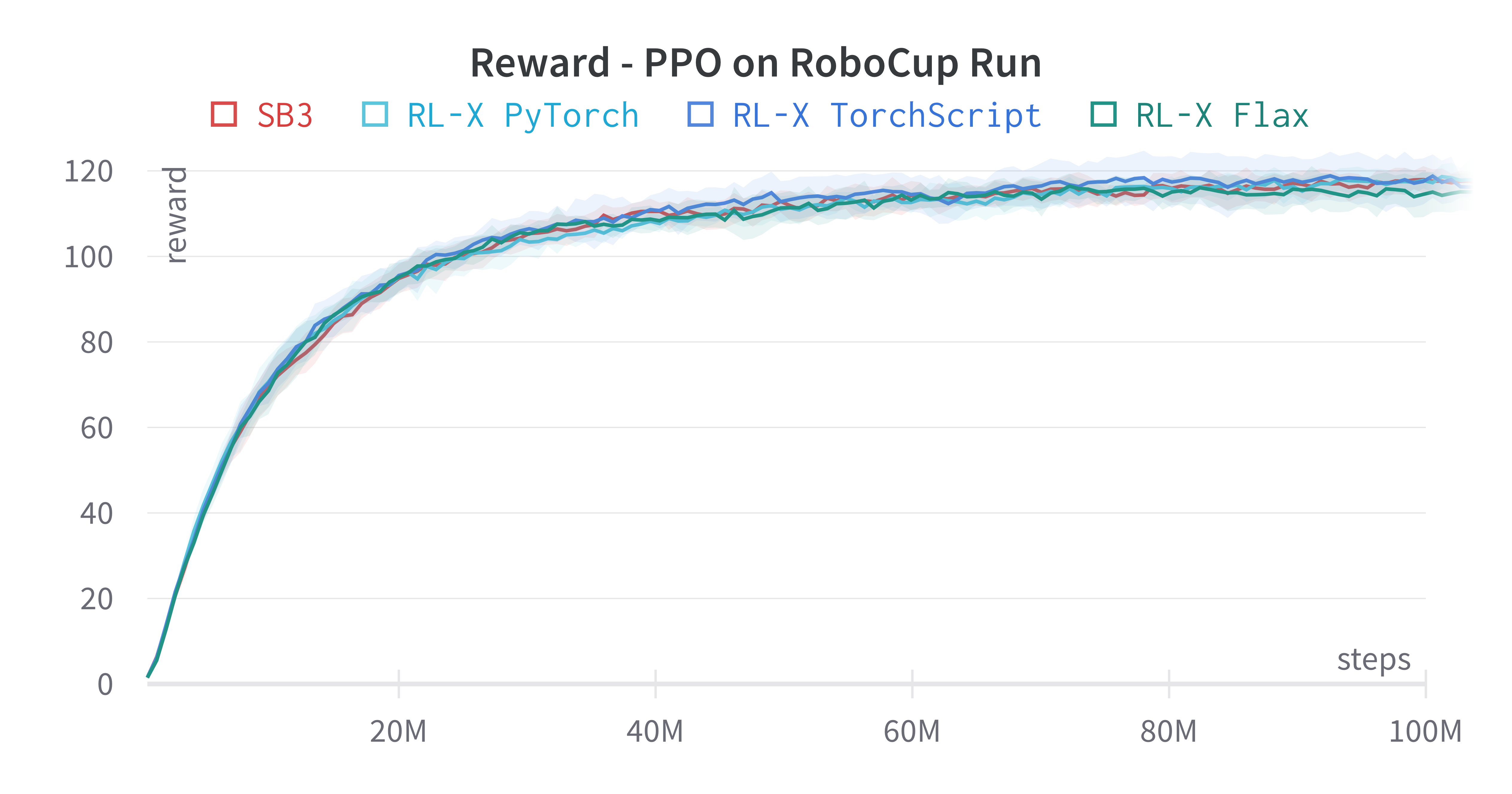}
    \end{center}
    \caption{Comparison of the reward performance in the RoboCup Run environment of the RL-X implementations for PPO in PyTorch, TorchScript and Flax to the SB3 baseline.}
    \label{fig:ppo_reward}
\end{figure}

\begin{figure}[htb]
    \begin{center}
        \includegraphics[width=\linewidth]{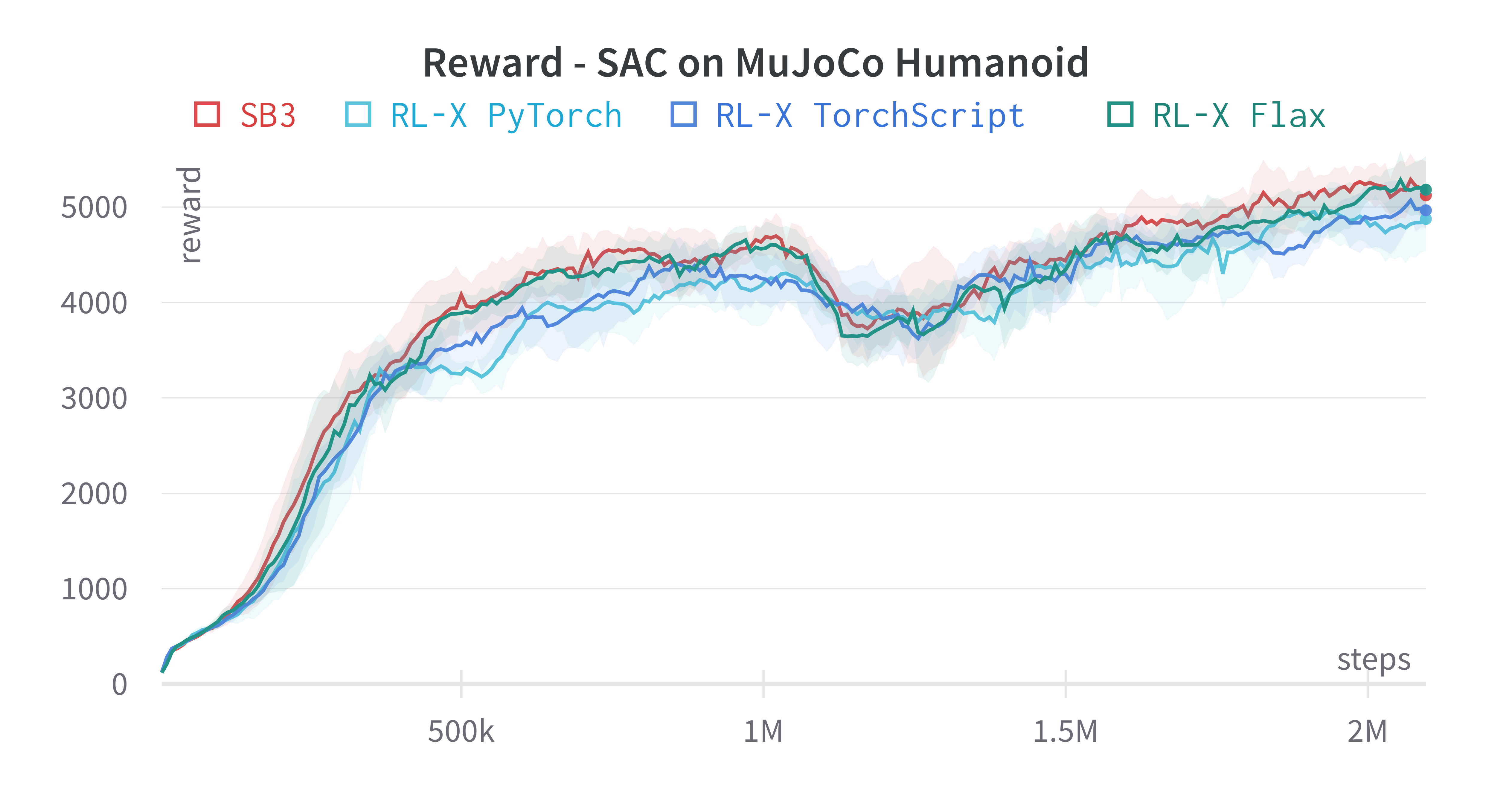}
    \end{center}
    \caption{Comparison of the reward performance in the MuJoCo Humanoid-v4 environment of the RL-X implementations for SAC in PyTorch, TorchScript and Flax to the SB3 baseline.}
    \label{fig:sac_reward}
\end{figure}

\clearpage

Figure~\ref{fig:ppo_reward} shows the reward performance of the PPO implementations in the RoboCup Run environment.
As expected, all four reward curves are very similar to another and show essentially the same performance on the task.

Figure~\ref{fig:sac_reward} shows the reward performance of the SAC implementations in the MuJoCo Humanoid-v4 environment.
Again, all four versions show very similar performance.
The minor differences can be attributed to the factors mentioned above and lay within the standard deviation of the reward curves.

\subsection{Computational Performance}
How fast a RL algorithm can be trained depends on the computational performance of the algorithm itself and the (simulation) speed of the environment.
We benchmark the relative computational performance of the RL-X implementations for PPO and SAC and compare them to the SB3 baseline.
The same environments are used as in the previous section and each version of the two algorithms is trained three times with different seeds.
The runtime of an algorithm can depend on the current policy of the agent, as more or less episode resets can be triggered, various actions can consume various processing time and different parts of the environment can be different in their computational complexity.
To limit the influence of those factors, the experiments only last for 500k steps with PPO and 50k steps with SAC to ensure that the agents policies don't diverge too much from each other.

Hardware components do naturally have a huge impact on the runtime.
Therefore each experiment in this section is run on the same hardware to ensure a fair comparison.
We used the following hardware setup:
\begin{itemize}
    \item CPU: Intel Core i9-9900K
    \item GPU: NVIDIA GeForce RTX 2080 Ti
    \item RAM: Corsair VENGEANCE 4x16GB, DDR4, 3200MHz
\end{itemize}

\begin{figure}[htb]
    \begin{center}
        \includegraphics[width=\linewidth]{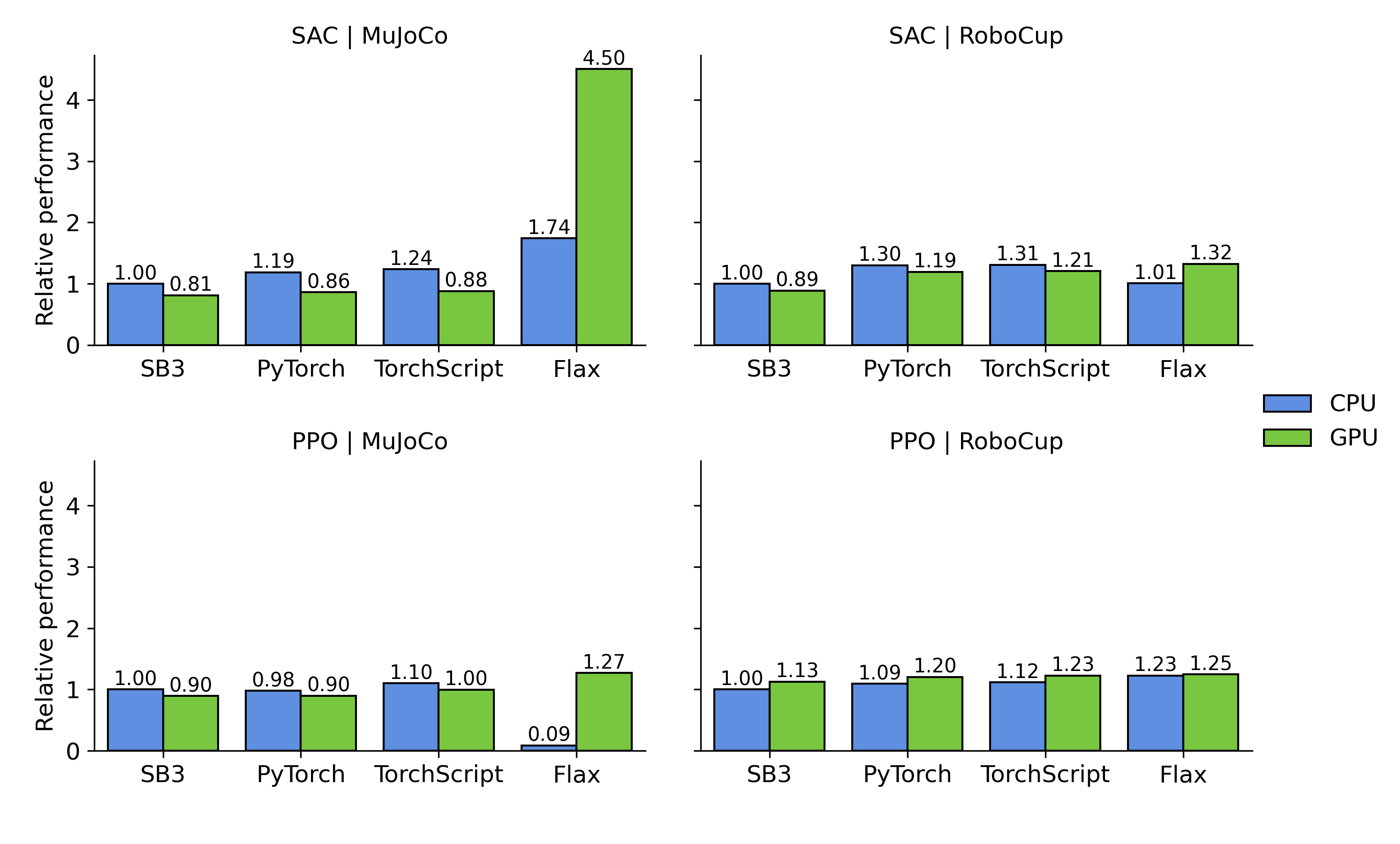}
    \end{center}
    \caption{Comparison of the relative computational performance of the RL-X implementations for PPO and SAC in PyTorch, TorchScript and Flax to the SB3 baseline.}
    \label{fig:bar_plot}
\end{figure}

Figure~\ref{fig:bar_plot} shows the measured relative performances of the different algorithm implementations and versions averaged over all timesteps and the three seeds.
The error bars for the standard deviation were omitted for visual clarity and for the fact that they were not exceeding more than +/- 1\% performance.
To calculate the plotted relative performance, the measured performances are normalized by the performance of the SB3 CPU baseline in the respective quadrant of the chart.

The top left bar chart shows the performance of the SAC implementations in the MuJoCo Humanoid-v4 environment.
This combination of algorithm and environment shows the most significant performance improvements through the RL-X implementations.
Compared to the RoboCup task, the MuJoCo environment is much less computationally demanding and resembles less of a computational bottleneck in the RL training loop.
A step in the MuJoCo environment takes around 0.001 seconds on average, which is 25 times less than the average step time of 0.025 seconds in the RoboCup environment.
This means that improvements in the computational performance of the algorithm are more noticeable.
Off-policy algorithms like SAC compute their loss and update their parameters in every timestep by sampling the used batch from a replay buffer, whereas on-policy algorithms like PPO only do so after a full batch of newly collected trajectories is available.
This optimization in every step is the reason why SAC undergoes a much bigger performance improvement than PPO.
PPO's performance is more dependent on the runtime of the environment and on over how many parallel agents it can distribute the data collection.

In the ideal case of using a fast simulated environment (here MuJoCo) and an off-policy algorithm (here SAC), Flax can speed up the training by a factor of 4.5x with the help of a GPU compared to SB3's PyTorch implementation running on CPU.
For the more infrequent updating PPO the speedup with Flax is not as drastic but still noticeable with a factor of 1.27 and 1.25 on the MuJoCo and RoboCup environments respectively.
In the case of the slower RoboCup environment, RL-X's PyTorch, TorchScript and Flax implementations achieve all roughly the same performance and are around 1.2-1.3x faster than SB3.
In all our experiments, the Flax GPU versions achieved the best computational performance, especially when the acting in the environment is not the bottleneck.
The TorchScript versions offer a consist boost in performance over the pure PyTorch implementations, which is great as it barely requires additional code changes.
Noticeable is that using the GPU with PyTorch and TorchScript had a negative impact on the performance, with the PPO + RoboCup setup being the exception.
This is most likely due to the small batch sizes we used.
In this regime the cost of copying the data to the GPU is higher than the performance gain from its fast processing capabilities.
On the contrary, Flax handles this better, as its GPU version outperforms the CPU version in all our experiments.
Unfortunately, the JIT compilation with JAX can suffer from high initial compilation times on the CPU.
This is not the case for TorchScript's JIT.
Additionally the poor performance of the Flax CPU version on the MuJoCo environment should be noted, but we have not yet found the reason for this.
Finally we do not yet have results on the performance of JAX on TPUs and leave this for future work.

In addition to the main results with SB3, a small set of comparisons with other frameworks, namely RLlib and CleanRL, are conducted in the SAC + PyTorch + GPU + MuJoCo setup.
RL-X's implementation shows a 5.1x speed up over RLlib's and the same performance as CleanRL's implementation.

\section{Conclusion and Future Work}
\label{conclusions}
In this paper we have presented a new Deep Reinforcement Learning framework RL-X, which has demonstrated major improvements over existing libraries in terms of runtime performance, implementation clarity and extensibility.
Our experiments conducted on a custom RoboCup Soccer Simulation 3D environment and the MuJoCo Humanoid-v4 environment have shown that RL-X achieves comparable training results to Stable-Baselines3 in terms of learning quality, as expected.
Notably, RL-X has showcased impressive walltime efficiency, up to 4.5 times faster in a classic SAC + MuJoCo setup running on GPU with the help of JAX and JIT compilation.
Overall, our results suggest that RL-X is a highly effective and efficient tool for DRL research and applications.

To ensure that RL-X stays up to date with the latest developments in RL, we plan to regularly add new algorithms and features to the library.
Especially newer state-of-the-art algorithms like Muesli \cite{hessel2021} and V-MPO \cite{song2019}, which completely lack open-source implementations, are on our roadmap.
Additionally, we will soon add support for offline RL datasets and implementations of algorithms in the field of intrinsic motivation, as RL-X is already in full use in multiple research projects in those areas.
A separate evaluation loop during training is the first feature extension we plan to add to RL-X.

While we provide the framework documentation and further literature on all algorithms in extensive readme files, we plan to further improve the documentation and add more examples and tutorials for our users.
More Python type hints and a broad test coverage are also on our list.

To enable further performance improvements, we want to benchmark RL-X on TPUs, as JAX was developed with this hardware in mind.
PyTorch 2.0 did also just release with a new JIT compile system, which might improve over TorchScript's JIT and would allow for a merge of both implementation versions.


\section*{Acknowledgement}
Thanks to the magmaOffenburg team for providing the tools used in this paper.

\addtolength{\textheight}{-2cm}   



\section*{Appendix}
\begin{table}[htb]
\caption{Hyperparameter settings used in the experiments}
\begin{center}
    \begin{tabular}{| p{5em} || p{10em} | p{4em} |}
    \hline
    Algorithm & Parameter & Value \\ \hline \hline
    \multirow[t]{10}{2em}{PPO} & clip\_range & 0.2 \\
    & critic\_coef & 0.5 \\
    & entropy\_coef & 0.0 \\
    & gae\_lambda & 0.95 \\
    & max\_grad\_norm & 0.5 \\
    & minibatch\_size & 1536 \\
    & nr\_epochs & 10 \\
    & nr\_steps & 2048 \\
    & std\_dev & 0.3 \\
    & environment.nr\_envs & 24 \\ \hline
    \multirow[t]{8}{2em}{SAC} & batch\_size & 256 \\
    & buffer\_size & 1000000 \\
    & learning\_starts & 5000 \\
    & log\_std\_min & -20 \\
    & log\_std\_max & 2 \\
    & target\_entropy & "auto" \\
    & tau & 0.005 \\
    & environment.nr\_envs & 1 \\ \hline
    \multirow[t]{4}{2em}{Both} & anneal\_learning\_rate & False \\
    & gamma & 0.99 \\
    & learning\_rate & 0.0003 \\
    & nr\_hidden\_units & 64 \\ \hline
    \end{tabular}
\end{center}
\label{tb:hyperparams}
\end{table}

\end{document}